\documentclass[sigconf]{acmart}

\usepackage{balance}

\AtBeginDocument{%
 }

\copyrightyear{2024}
\acmYear{2024}
\setcopyright{rightsretained}
\acmConference[WWW '24 Companion]{Companion Proceedings of the ACM Web Conference 2024}{May 13--17, 2024}{Singapore, Singapore}
\acmBooktitle{Companion Proceedings of the ACM Web Conference 2024 (WWW '24 Companion), May 13--17, 2024, Singapore, Singapore}
\acmDOI{10.1145/3589335.3651909}
\acmISBN{979-8-4007-0172-6/24/05}

\makeatletter
\gdef\@copyrightpermission{
  \begin{minipage}{0.3\columnwidth}
   \href{https://eur05.safelinks.protection.outlook.com/?url=https://creativecommons.org/licenses/by/4.0/&data=05|02|D20125664@mytudublin.ie|a5390306b9b3402d1be208dc490501a5|766317cbe9484e5f8cecdabc8e2fd5da|0|0|638465533530864684|Unknown|TWFpbGZsb3d8eyJWIjoiMC4wLjAwMDAiLCJQIjoiV2luMzIiLCJBTiI6Ik1haWwiLCJXVCI6Mn0=|0|||&sdata=9e/Q0UjYcp0nUhTOJGm7JnMa/QMicURTqbUAIT3BLVU=&reserved=0}{\includegraphics[width=0.90\textwidth]{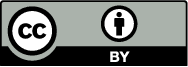}}
  \end{minipage}
  \hfill
  \begin{minipage}{0.7\columnwidth}
   \href{https://eur05.safelinks.protection.outlook.com/?url=https://creativecommons.org/licenses/by/4.0/&data=05|02|D20125664@mytudublin.ie|a5390306b9b3402d1be208dc490501a5|766317cbe9484e5f8cecdabc8e2fd5da|0|0|638465533530870118|Unknown|TWFpbGZsb3d8eyJWIjoiMC4wLjAwMDAiLCJQIjoiV2luMzIiLCJBTiI6Ik1haWwiLCJXVCI6Mn0=|0|||&sdata=MeROvfVgut5m5J3G7CWaw3htk17XMjLXbtFCyg6o3ik=&reserved=0}{This work is licensed under a Creative Commons Attribution International 4.0 License.}
  \end{minipage}
  \vspace{5pt}
}
\makeatother

\begin{document}

\title[GPT Assisted Annotation]{GPT Assisted Annotation of Rhetorical and Linguistic Features for Interpretable Propaganda Technique Detection in News Text.}

\author{Kyle Hamilton}
\email{kyle.i.hamilton@mytudublin.ie}
\orcid{0000-0003-0809-0664}
\affiliation{%
  \department{School of Computer Science and SFI Centre for Research Training in Machine Learning}
  \institution{Technological University Dublin}
  \city{Dublin}
  \country{Ireland}
}

\author{Luca Longo}
\email{luca.longo@tudublin.ie}
\orcid{0000-0002-2718-5426}
\affiliation{%
  \department{Artificial Intelligence and Cognitive Load Research Lab, School of Computer Science and SFI Centre for Research Training in Machine Learning}
  \institution{Technological University Dublin}
  \city{Dublin}
  \country{Ireland}
}

\author{Bojan Bo{\v{z}}i{\'c}}
\email{bojan.bozic@tudublin.ie}
\orcid{0000-0002-4420-1029}
\affiliation{%
  \department{School of Computer Science and SFI Centre for Research Training in Machine Learning}
  \institution{Technological University Dublin}
  \city{Dublin}
  \country{Ireland}
}

\renewcommand{\shortauthors}{Kyle Hamilton, Luca Longo, \& Bojan Božić}

\begin{abstract}
While the use of machine learning for the detection of propaganda techniques in text has garnered considerable attention, most approaches focus on ``black-box'' solutions with opaque inner workings. Interpretable approaches provide a solution, however, they depend on careful feature engineering and costly expert annotated data. Additionally, language features specific to propagandistic text are generally the focus of rhetoricians or linguists, and there is no data set labeled with such features suitable for machine learning. This study codifies 22 rhetorical and linguistic features identified in literature related to the language of persuasion for the purpose of annotating an existing data set labeled with propaganda techniques. To help human experts annotate natural language sentences with these features, RhetAnn, a web application, was specifically designed to minimize an otherwise considerable mental effort. Finally, a small set of annotated data was used to fine-tune GPT-3.5, a generative large language model (LLM), to annotate the remaining data while optimizing for financial cost and classification accuracy. This study demonstrates how combining a small number of human annotated examples with GPT can be an effective strategy for scaling the annotation process at a fraction of the cost of traditional annotation relying solely on human experts. The results are on par with the best performing model at the time of writing, namely GPT-4, at 10x less the cost. Our contribution is a set of features, their properties, definitions, and examples in a machine-readable format, along with the code for RhetAnn and the GPT prompts and fine-tuning procedures for advancing state-of-the-art interpretable propaganda technique detection.
\end{abstract}




\maketitle

\section{Introduction}

The study of rhetoric has a rich history spanning thousands of years, traditionally falling within the purview of the humanities and relying on manual analysis methods. Given the limitations in the scalability of manual analysis, we naturally turn to exploiting machine learning approaches. In turn, the challenge for the machine learning community is a shortage of labelled data. Hence, most machine learning-based approaches have been executed on small data sets annotated with specific rhetorical figures, such as, for example, ``chiasmus'' \cite{Dubremetz_Nivre_2018}, ``litotes'' \cite{Mitrović_O’Reilly_Harris_Granitzer_2020}, or ``doubt'' \cite{Ward_Link_Avramov_Goodwin_2022}, to name a few. 

Both propagandistic techniques and rhetorical devices can be considered tools of persuasion and have certain overlapping characteristics. Indeed, ``the major techniques of propaganda are long known rhetorical techniques gone wrong'' \cite{Bryant_1953}. To our knowledge, the most relevant data set related to rhetoric in news text is the Propaganda Technique Classification (PTC) corpus developed by \citet{Da_San_Martino_2019}. The PTC corpus is annotated with 18 propaganda techniques divided into ``logical fallacies'' and ``emotion and sentiment mechanisms''. These can be considered logos and pathos in the nomenclature of rhetoric.\footnote{Notably, ``appeals to authority'' is categorized under ``logical fallacies'', whereas, in the rhetoric tradition, this would belong to a third category, namely ``ethos''. } Furthermore, some propaganda techniques, such as ``doubt'', have equivalent counterparts in rhetorical devices. In contrast, others tend to have a broader scope, such as ``flag-waving,'' which includes appeals to nationalism as a means of persuasion. 

We propose that propaganda techniques and rhetorical devices can be further deconstructed into grammatical and word choice components as developed by \citet{fahnestock2011rhetorical} in \textit{Rhetorical Style: The Uses of Language in Persuasion}. If true, these components, or ``features'' as we will refer to them from here on in, should correlate with propaganda techniques. However, to test this assumption, we need to be able to extract the features from natural language in order to use them for model training and inference. No such extraction procedure exists for these specific features. Thus, the focus of this study is on constructing such a procedure. 

\begin{figure*}[htp]
    \centering
    \includegraphics[scale=0.65]{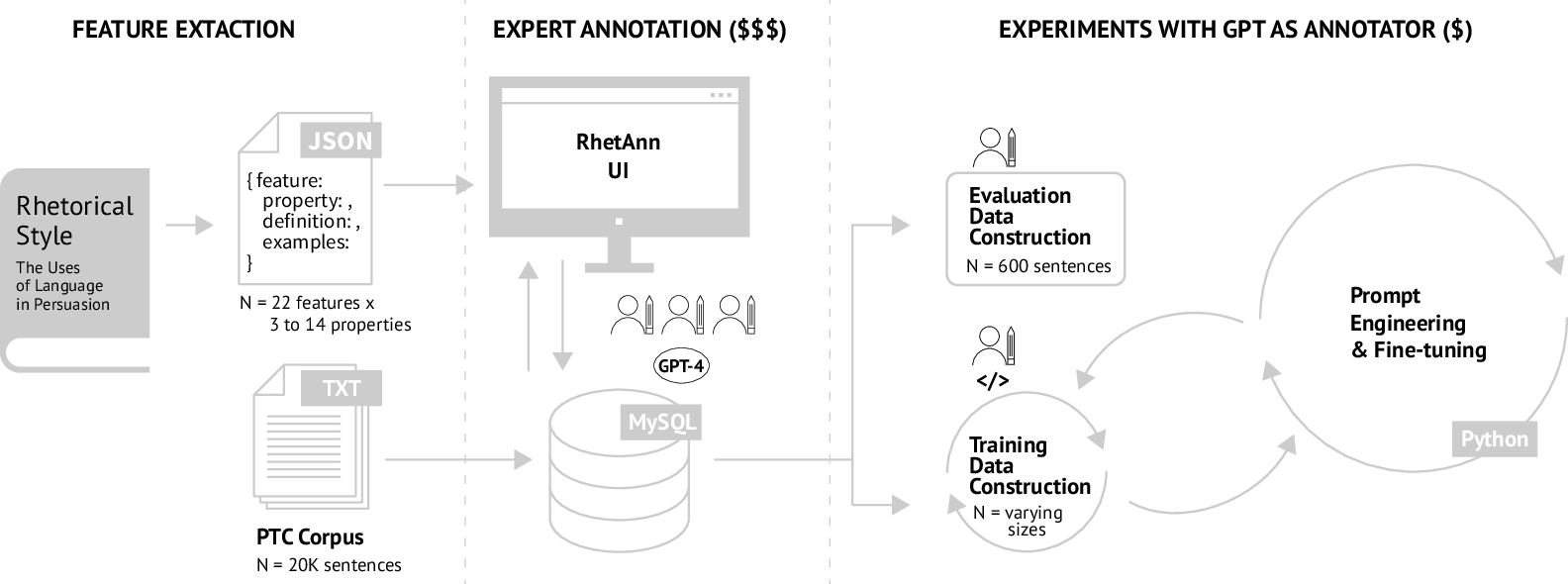}
    \caption{Three stages of GPT assisted annotation of rhetorical and linguistic features for interpretable propaganda technique detection. Code available in: \url{https://github.com/kyleiwaniec/rhetoric-annotation} }
    \label{fig:diagram}
\end{figure*}

We were further constrained by a very limited budget. This is not an unusual position to be in - cost-saving techniques can be critical to a researcher's ability to conduct their work. We propose an annotation strategy which takes advantage of GPT-4's explanations to aid in the human annotation phase, where a small subset of the data is annotated, and utilizes an iterative prompt engineering approach to annotate the remaining data using GPT-3.5. We show that even noisy human-labeled data can be utilized to fine-tune GPT-3.5 to achieve the performance of its state-of-the-art but 10x more expensive counterpart, GPT-4. Our process is illustrated in Figure \ref{fig:diagram}.

Even though our long-term goal is specific to propaganda detection, we believe that the features and annotation procedure we propose can be used in any NLP task that aims to leverage the characteristics of the language of persuasion. In addition, this effort could contribute to developing a linguistic rhetorical representation akin to existing representations such as Abstract Meaning Representation (AMR) or Rhetorical Structure Theory (RST).

\section{Methodology}\label{methodology}
\subsection{Data}

\begin{figure}[htp]
    \centering
    \includegraphics[scale=0.2]{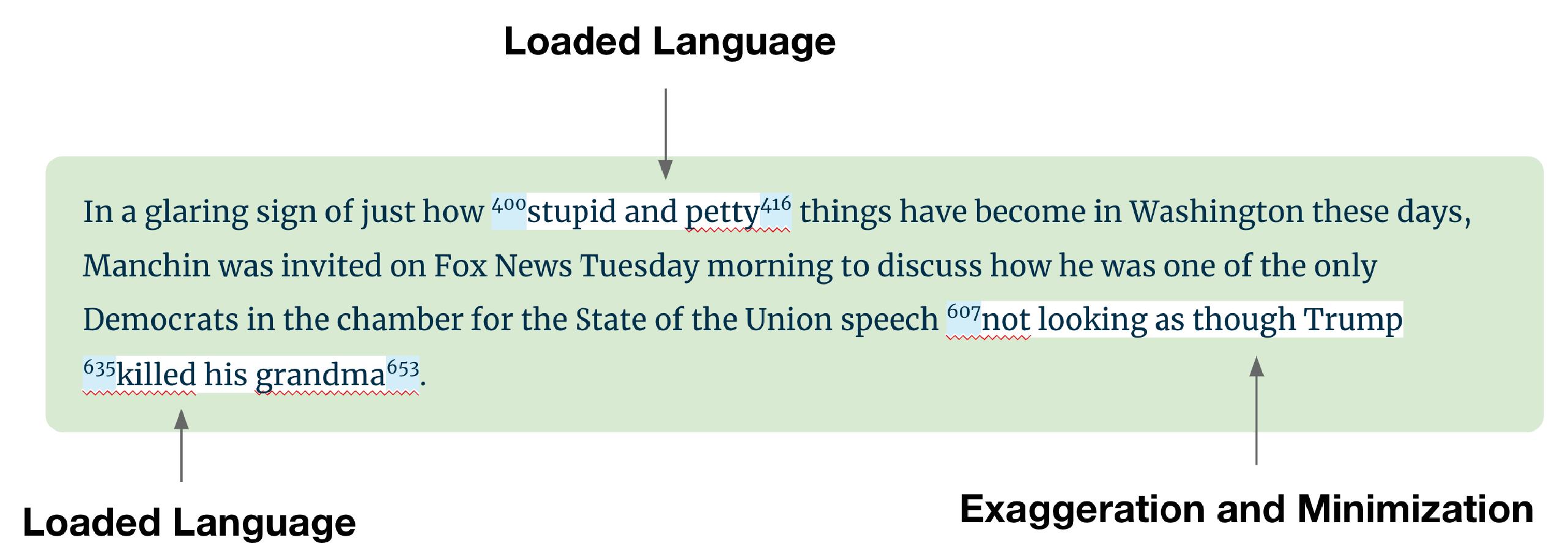}
    \caption{A sentence from the PTC corpus annotated at the fragment level with three occurrences of propaganda techniques. Fragments are highlighted in white. }
    \label{fig:fragment}
\end{figure}

We utilize the PTC corpus, introduced at the NLP4IF workshop at EMNLP-IJCNLP 2019, which consists of 451 news articles annotated with 18 propaganda techniques at the sub-sentence fragment level. This means a sentence can have multiple fragments labelled with multiple propaganda techniques. Figure \ref{fig:fragment} illustrates an example sentence from PTC. Details about this data set and associated tasks can be found in \citet{Da_San_Martino_2019}. We roll up the labels to the sentence level as per \citet{Yu_Martino_Mohtarami_Glass_Nakov_2021}, which, at the time of this writing, represents the state-of-the-art interpretable propaganda technique detection in news text. Thus, fragment location information is discarded, and each sentence is labelled with zero or more propaganda techniques. 
This aligns with similar open problems devoted to creating interpretation techniques for opaque, black-box generative and large language models \cite{LONGO2024102301}, both at the global and local levels.

\subsection{Features}
We use \citet{fahnestock2011rhetorical} as a manual to codify 22 features, each with anywhere from 3 to 14 possible properties, summarized in Table \ref{table:features} of the appendix. 
\citet{fahnestock2011rhetorical}'s book is organized into four parts: I) Word Choice, II) Sentences, III) Interactive dimensions, and IV) Passage construction. We utilize only parts I and II as they can be applied at the sentence level and leave III and IV for future work. For each property we generate a brief definition by summarizing the corresponding section in the book and extracting one or more examples. See Table \ref{tab:hyperbole} for the \textit{hyperbole} property of the \textit{Tropes} feature. 

\begin{table}[t]
\small
\begin{center}
\begin{tabular}{p{3.2in}}
\bf I. Word Choice\\ 
\toprule
Tropes \\
\midrule
hyperbole - An overstatement. An exaggerated statement or claim not meant to be taken literally. Example: ``I'm so hungry I could eat a horse.''\\
\bottomrule
\end{tabular}
\end{center}
\caption{\label{tab:hyperbole}Definition and example of the \textit{hyperbole} property of the \textit{Tropes} feature.}
\end{table}

\subsection{Human annotations}

We engaged a professional annotation firm to annotate a subset of the PTC corpus with our features. This task was undertaken by three independent human experts, each with a background in linguistics. 

To estimate the task's difficulty, we annotated 15 sentences, randomly selected from the PTC corpus, using a spreadsheet where each sentence is a row, and each feature is a column with drop-down menus containing the properties. Definitions and examples were stored in separate sheets for reference as needed. This process proved cumbersome and time-consuming as it required multiple clicks, moving back and forth between sheets, finding the relevant row, and scrolling to the appropriate column at each move from one sheet to another. With so many redundant actions and the necessity of keeping track of the current sentence and feature in memory, completing the annotation process for one thousand sentences was estimated at 1 year, working 8 hours a day, 40 hours a week. To reduce the time needed to complete the task, we built a web-based user interface called RhetAnn.

\begin{figure*}
    \centering
    \includegraphics[scale=0.35]{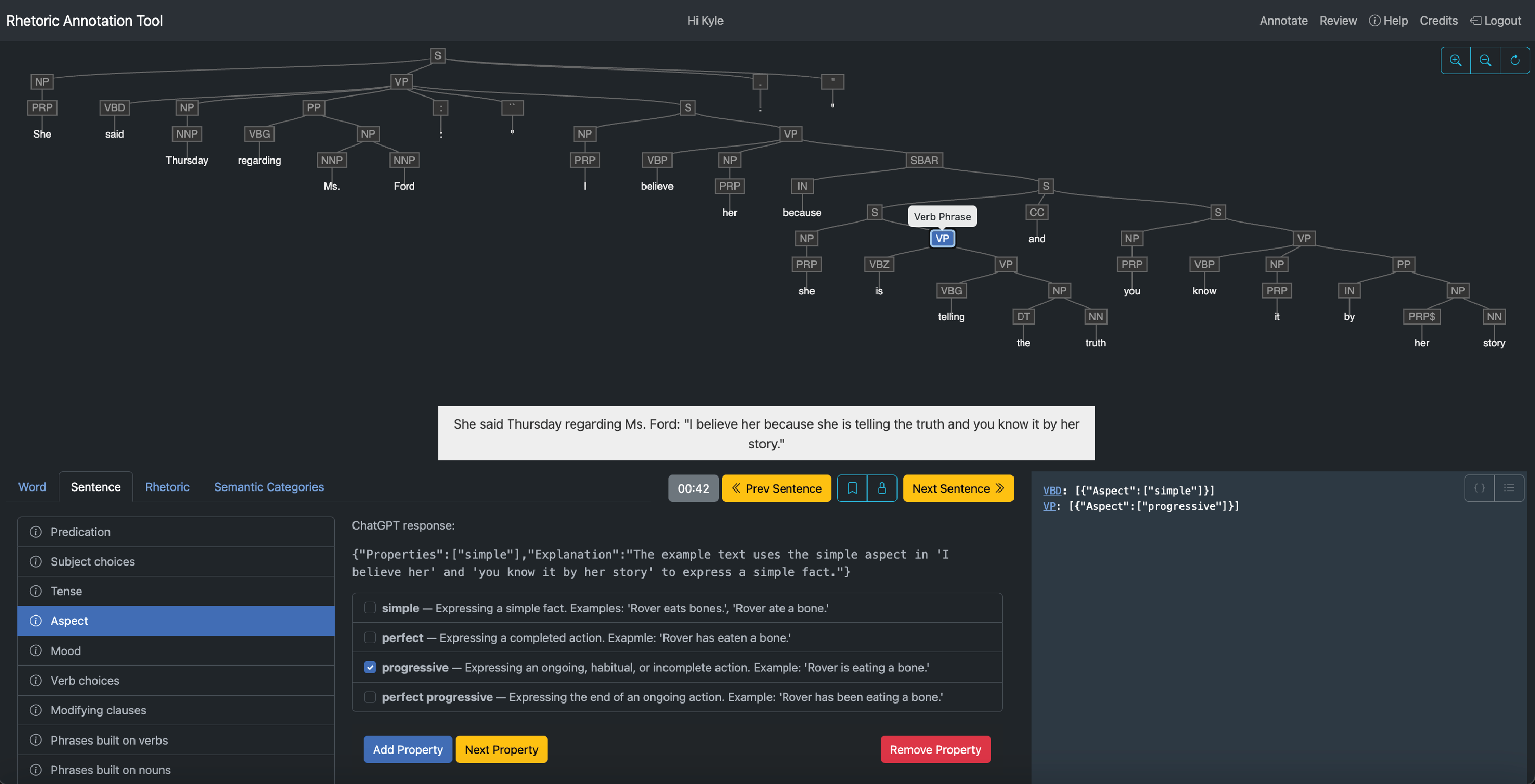}
    \caption{RhetAnn - Rhetoric Annotation web-based application. The \textit{Verb Phrase} (VP) node is selected in the parse tree. The current feature is \textit{Aspect}, and the selected property is \textit{progressive}.}
    \label{fig:rhetann}
\end{figure*}

\subsection{Annotation user interface – RhetAnn }
The RhetAnn user interface - Figure \ref{fig:rhetann} - contains all of the information needed for a given sentence and feature, including all property definitions and examples, and allows the annotator to move easily between features and sentences without having to keep anything in memory. In addition to reducing the complexity associated with the above-mentioned actions, we have added a GPT-4 assistant. During our initial estimation efforts, we found that the explanations provided by GPT-4 were sometimes helpful and could speed up the annotation process. This is similar to the finding in \citet{Wei_Wang_Schuurmans_Bosma_Ichter_Xia_Chi_Le_Zhou_2023}. Annotators were instructed to use the assistant only as an aid and that the final annotation must always be verified. The prompt to GPT-4 consists of the same instructions and definitions/examples as given to the human annotators. A sample prompt is given in Table \ref{tab:prompt}. Each time an annotator submits an annotation, the GPT-4 API is called, and the response is stored.

\begin{table}[t]
\small
\begin{center}
\begin{tabular}{p{3.2in}}
\bf Prompt\\ 
\toprule
You are a rhetorician and linguist specializing in news text. \\
Your task is to identify which, if any, of the following properties of Aspect are used in the example text.   You may select multiple properties. Each line contains a property followed by a colon, followed by a brief definition and example(s):\\
simple: Expressing a simple fact. Examples: 'Rover eats bones.', 'Rover ate a bone.'\\
perfect: Expressing a completed action. Example: 'Rover has eaten a bone.'\\
progressive: Expressing an ongoing, habitual, or incomplete action. Example: 'Rover is eating a bone.'\\
perfect progressive: Expressing the end of an ongoing action. Example: 'Rover has been eating a bone.'\\
Format your response as a JSON object with ``Properties'' and ``Explanation'' as the keys.  The value of ``Properties'' should be a list. If none of the properties are present, return an empty list.  Explain your choice in the ``Explanation''. Make your response as short as possible. The example text is delimited with triple backticks. \\
Example text: \verb|```|While it earlier made attempts to balance its shoddier side with some interesting reporting, it is now solidly mainstream in the worst sense.\verb|```|\\
\bottomrule
\end{tabular}
\end{center}
\caption{\label{tab:prompt}Example prompt posted to GPT-4 API for the \textit{Aspect} feature.}
\end{table}

For properties that apply to single words or sentence fragments, we want our annotators to select the appropriate part of the sentence along with the property. To that end, we provide a constituency parse tree for the annotator to select a node which contains the desired fragment, illustrated in Figure \ref{fig:rhetann}. To generate the parse trees, we utilise the SpaCy python library, which is open source and makes data preparation easily reproducible. Annotating the data at this level of granularity creates opportunities for a wide range of NLP tasks which aim to enrich such parse trees for downstream tasks, and in particular, for architectures involving graphs \cite{Liu_Wu_2022,Wu_Chen_Shen_Guo_Gao_Li_Pei_Long_2023}. The parse tree structure does not always allow the annotator to capture the fragment exactly as they would have done if this limitation wasn't imposed. However, utilizing parse trees will make it easier to use our data to complement research on language representations, which also utilize this structure, which we believe justifies the above limitation. 

To initiate the annotation process, we invited the annotation firm to a tutorial on the UI, data, and labels. Next, we had three annotators independently annotate the first 10 sentences. We then reconvened in a virtual meeting to address inconsistencies and answer any clarifying questions that may have come up in this phase. Most of the inconsistencies involved assigning nodes in the parse tree. Some annotators tended to assign nodes closer to the leaves, while others chose nodes higher up the tree, thus selecting a larger sentence fragment. We resolved this by instructing the annotators to choose nodes as close as possible to the leaves, or in other words, the smallest possible fragment that captures the feature/property under consideration. Inconsistencies arising from different annotators' interpretations were discussed until consensus was reached to help annotators align their understanding moving forward. We discuss the inter-annotator agreement in detail in Section \ref{promt-engineering}. Annotators were also encouraged to refer to the \citet{fahnestock2011rhetorical} book for additional clarification.  

Our budget necessitated a trade-off between quantity and reliability. Using our alignment exercise described above as an indicator, we anticipated that inter-annotator agreement would be low, and as such, we would have preferred to have more annotators. However, this would have significantly reduced the number of sentences we could have annotated. We believe that three annotators are an acceptable minimum in our case, and thus, the number of sentences is driven by this assumption. Consequently, we chose to have three annotators independently annotate 350 sentences. We took a stratified sample balanced across the propaganda techniques in the full training data set.
As there are three annotators, GPT-4 will be called three times for each sentence/feature combination, adding to our cost calculation. For this small subset of sentences, we chose to use the most powerful zero-shot GPT-4 model available at the time of writing. To annotate the remainder of the data, we will utilise the orders of magnitude less expensive GPT-3.5-turbo model.

\subsection{Inter-annotator agreement}\label{inter-annotator-agreement}

\begin{figure}[htp]
    \centering
    \includegraphics[scale=0.9]{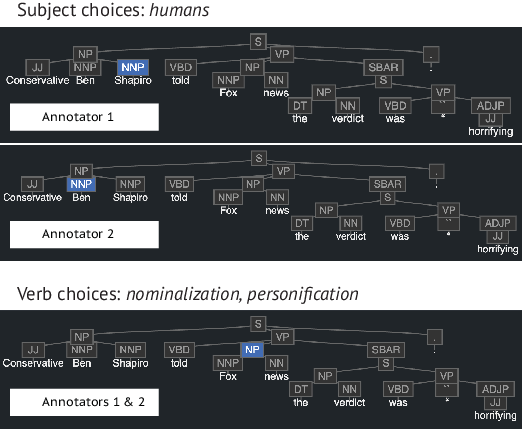}
    \caption{For the \textit{Subject choices} feature two annotators assign \textit{humans} but select different nodes. For the \textit{Verb choices} feature, one annotator assigns \textit{nominalization}, while the other assigns \textit{personification}. We consider the first case in agreement and the second one not.}
    \label{fig:agreement-example}
\end{figure}

Each sentence can be annotated with one or more of 22 features, where each feature can have between 3 and 14 properties. Annotators can assign multiple properties for a given feature. This makes the space of possibilities very large, and the expectation of all annotators assigning the same combination of features/properties and selecting the same node in the parse tree would be unreasonable. Instead, we measure agreement on a per-feature basis, and we disregard the node selection. An illustrated example is given in Figure \ref{fig:agreement-example}. In this example, both annotators assign \textit{humans} for the \textit{Subject choices} feature but select different nodes in the parse tree. For the \textit{Verb choices} feature, one annotator assigns \textit{nominalization}, while the other assigns \textit{personification}. We consider \textit{Subject choices} to be in agreement and \textit{Verb choices} not in agreement.

We calculate Krippendorff's alpha for each feature among all three annotators. Krippendorff's alpha is a statistical measure of inter-coder agreement which considers disagreement one would expect to get by chance. Features which do not have any labels assigned by any of the annotators are ignored. It's important to note that each feature can have only one label in the Krippendorff statistic. Since our setting allows for multiple labels, we cannot use this metric to measure partial agreement. For example, an annotator who assigns both \textit{negation} and \textit{modality} to \textit{Verb choices} is considered in disagreement with an annotator who only assigns \textit{modality}. In order to address this scenario, we also calculated the Jaccard score for each feature. This gives us a measure of partial agreement. We sum these scores and normalize them by the number of examples for the given feature. Not all features have the same number of examples, as not all sentences are annotated with all features. For example, the \textit{Tropes} feature may not occur in all 350 sentences because not all sentences make use of tropes. 


For completeness, we also calculate the joint probabilities of exact agreement, and again, we normalize by the number of examples for the given feature. We can use these measures to compare the inter-annotator agreement to intra-GPT consistency, the extent to which the GPT output is consistent given the same input. The GPT APIs are called three times for each prompt. Consistency is considered exact when the set of properties for a given feature is exactly the same in all three responses. For example, given the feature \textit{Verb choices}, response 1 = \{\textit{negation}, \textit{modality}\}, response 2 = \{\textit{negation}, \textit{modality}\}, and response 3 = \{\textit{negation}, \textit{modality}\}. All scores are shown in Table \ref{table:agreemet}.


\begin{table}[t]
\small
\begin{center}
\begin{tabular}{l|l|l|l|l|l}
\toprule
\bf Feature & \bf K. & \bf J. & \bf E. & \bf GPT-4 & \bf GPT-3.5\\ 
\midrule
Aspect & 0.364 & 0.573 & 0.320 & 0.980 & 0.568 \\
Emphasis & -0.023 & 0.325 & 0.314 & 0.588 & 0.389 \\
Fig. of argument & 0.330 & 0.101 & 0.094 & 0.925 & 0.429 \\
Fig. of word choice & 0.171 & 0.067 & 0.047 & 0.873 & 0.342 \\
Language of origin & -0.101 & 0.282 & 0.042 & 0.421 & 0.353 \\
Language varieties & 0.074 & 0.613 & 0.126 & 0.059 & 0.157 \\
Lexical fields & 0.057 & 0.269 & 0.036 & 0.364 & 0.129 \\
Modifying clauses & 0.496 & 0.355 & 0.255 & 0.759 & 0.322 \\
Modifying phrases & 0.294 & 0.543 & 0.370 & 0.160 & 0.375 \\
Mood & 0.458 & 0.870 & 0.846 & 0.935 & 0.734 \\
New words & 0.285 & 0.203 & 0.094 & 0.866 & 0.241 \\
Parallelism & 0.035 & 0.009 & 0.009 & 0.957 & 0.384 \\
Phrases b.o. nouns & 0.520 & 0.250 & 0.237 & 0.953 & 0.476 \\
Phrases b.o. verbs & 0.240 & 0.104 & 0.095 & 0.976 & 0.431 \\
Predication & 0.331 & 0.556 & 0.324 & 0.769 & 0.459 \\
Prosody \& punct. & -0.003 & 0.000 & 0.000 & 0.000 & 0.067 \\
Sentence arch. & 0.197 & 0.183 & 0.106 & 0.575 & 0.412 \\
Series & 0.332 & 0.089 & 0.089 & 1.000 & 0.412 \\
Subject choices & 0.351 & 0.426 & 0.245 & 0.652 & 0.331 \\
Tense & 0.477 & 0.648 & 0.447 & 0.900 & 0.353 \\
Tropes & 0.217 & 0.107 & 0.067 & 0.853 & 0.255 \\
Verb choices & 0.407 & 0.281 & 0.182 & 0.678 & 0.476 \\
\bottomrule
\end{tabular}
\end{center}
\caption{\label{table:agreemet} Inter-annotator agreement scores for all features, where K=Krippendorff's alpha, J=Jaccard, E=Joint probability of exact agreement, GPT-4=Joint probability of intra-GPT-4 consistency, and GPT-3.5=Joint probability of intra-GPT-3.5 consistency.}
\end{table}

\begin{figure}[htp]
    \centering
    \includegraphics[scale=0.8]{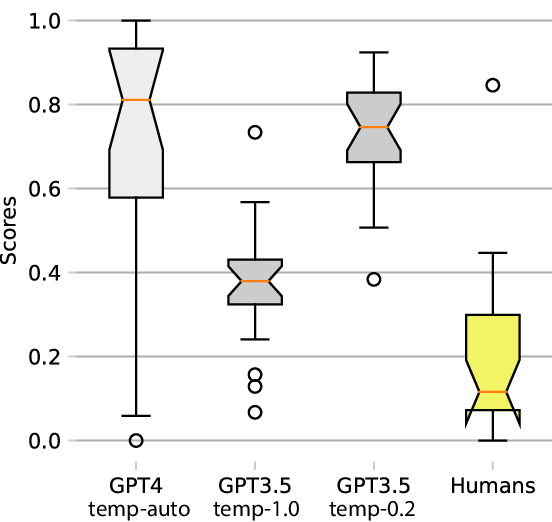}
    \caption{Distribution of intra-GPT/inter-annotator consistency scores across all features. GPT-3.5 was parameterized with a temperature setting of 1.0 and 0.2.}
    \label{fig:gpt-vs-human}
\end{figure}

\section{Experiments \& Results}

Interestingly, both intra-GPT-3.5 and intra-GPT-4 consistency is higher than inter-annotator agreement, illustrated in Figure \ref{fig:gpt-vs-human}. Consistency alone, however, does not tell us anything about accuracy. To measure accuracy, we sample 30 examples in each feature where all three annotators agree on the same list of properties and manually produce ground truth labels for those examples. Human accuracy, excluding a few exceptions, ranges from 70-100\%, whereas GPT-3.5 performs very poorly. GPT-4 performs five times better than GPT-3.5 but at ten times the cost. The results are shown in Table \ref{table:results} in the \textit{V1} columns. In addition, reducing the temperature parameter\footnote{\url{https://platform.openai.com/docs/api-reference/chat/create\#chat-create-temperature}} significantly increases the consistency of GPT-3.5. This is as expected. Notably, consistency in the GPT responses is not always correlated with accuracy. Thus, simply taking a majority vote is not a useful strategy in our case. To improve accuracy, we perform the following experiments.

\subsection{Prompt engineering}\label{promt-engineering}


Our initial zero-shot prompts, which were simply the same instructions provided to the human annotators, are ineffective. We asked GPT to provide a list of applicable properties in its response and an explanation of its choices. We scrutinize each of the approximately 600 (30 sentences * 22 features) GPT-generated explanations to understand where GPT is going wrong and thereby aim to improve the prompts. We categorise the GPT errors into the following types: 

\begin{itemize}
    \item \textbf{confounding} - GPT returns some property but uses a definition of a different property in its explanation.
    \item \textbf{over-generalizing} - GPT returns some property but makes up its own definition of that property rather than the definition provided. Usually, this GPT definition is broader than the one provided and is out of context. This is despite being instructed to use the definitions and context provided.
    \item \textbf{hallucinating} -  GPT returns an explanation based on elements not present in the given sentence.
    \item \textbf{greedy answering} - GPT returns properties when none of the properties are applicable. This is despite being instructed to return an empty list in such cases.
    \item \textbf{other errors} - GPT returns properties and explanations that are not plausible or are simply incorrect.
\end{itemize}

We can solve for confounding and greedy answering by modifying the prompts to ask about one property at a time. Instead of providing a list of properties and definitions for GPT to choose from, we provide a single property and its definition and ask GPT if it is present. For each property we ask for a yes/no answer and an explanation. Given that majority voting does not improve accuracy, we set the temperature to 0.0 for maximum consistency and only prompt GPT once to reduce cost. This modification improves performance of GPT-3.5 on some features but does not change the overall performance of GPT-4. Results are listed in Table \ref{table:results} in the \textit{V2} columns. We do not anticipate any significant improvements with further prompt engineering as per \citet{Webson_Pavlick_2022}, who show that ``models learn just as fast with many prompts that are intentionally irrelevant or even pathologically misleading as they do with instructively `good' prompts. [...] these results contradict a hypothesis commonly assumed in the literature that prompts serve as semantically meaningful task instructions.'' That is to say, LLMs do not ``understand'' instructions in the same way that people do. Thus, prompt engineering efforts should focus on deconstructing the LLM rather than writing ``better'' instructions as perceived by humans. How to deconstruct the LLM is outside the scope of this study.

\subsection{Fine-tuning}\label{fine-tuning}

We construct three types of data sets for fine-tuning GPT-3.5 as follows: 
\begin{enumerate}
    \item \textbf{Small size, high quality, single data-set} - For each property of a given feature, we find one positive and one negative example from the sub-set of the data where we had previously generated ground truth labels. The pairings comprise the prompt (SYSTEM and USER message as per V2) and an answer (ASSISTANT message), a JSON object with `Answer' and `Explanation' keys and their values. This is the most labour-intensive data set as it requires verifying that the answer is correct and providing an explanation. 
    \item \textbf{Medium size, medium/high quality, single data-set} - For each property of a given feature we find 25 positive and 25 negative examples from the sub-set of the data which was annotated by humans. The pairings comprise the prompt (SYSTEM and USER message as per V2), and an answer (ASSISTANT message) which is a JSON object. We do not include an explanation. While omitting the explanations reduces the labour intensity, the number of examples nearly cancels this benefit. 
    \item \textbf{Large size, low quality, multiple data sets} - We utilize all human-annotated data minus the examples for which we generated ground-truth labels. For each sentence, feature and property, we generate positive examples by taking the majority vote of the human annotators and negative examples using sentences where none of the annotators applied the property under consideration. Each feature ends up with its own data set of size approx. 300 x \#properties. For example, the \textit{Aspect} feature has 4 possible properties, thus the size of this data-set is approx. 1,200 examples. This is the least labour-intensive effort as it is completely automated. The downside is that the majority vote is not always correct.
\end{enumerate}

To keep costs and labour down, we experiment with 2-4 features before committing to generating the full data sets. We fine-tune GPT-3.5 using the OpenAI web interface and evaluate the results on the held-out ground-truth data. The small and medium data sets produce rather poor results. This runs contrary to the recommendation from OpenAI of 50-100 high-quality examples.\footnote{\url{https://platform.openai.com/docs/guides/fine-tuning/preparing-your-dataset}} We abandon these experiments to avoid incurring unnecessary costs. The accuracy of the fine-tuned GPT-3.5 on the large, albeit noisy data sets, is, overall, on par with GPT-4, and even slightly higher for some features. Results are shown in the FT column of Table \ref{table:results}.


\begin{table}[t]
\small
\begin{center}
\begin{tabular}{l|l|l|l|l|l|l}
\multicolumn{2}{l}{{\bf }} & \multicolumn{2}{|l}{{\bf GPT V1}} & \multicolumn{2}{|l|}{{\bf GPT V2}} & \bf FT\\ 
\toprule
\bf Feature & \bf H. & \bf 3.5 & \bf 4 & \bf 3.5 & \bf 4 & \bf 3.5 \\ 
\midrule
Aspect & 0.94 & 0.14 & 0.46 & 0.31 & 0.44 & \textbf{0.5} \\
Emphasis & 0.6 & \textbf{0.4} & --  & 0.0 & 0.2 & 0.13 \\
Fig. of argument & 0.83 & 0.17 & \textbf{0.67}  & 0.53 & 0.57 & 0.47 \\
Fig. of word choice & 0.3 & 0.07 & \textbf{0.4}  & 0.03 & 0.17 & 0.37 \\
Language of origin* & -- & -- & --  & -- & -- & -- \\
Language varieties & 0.74 & 0.01 & 0.54  & 0.04 & 0.37 & \textbf{0.7} \\
Lexical fields & 0.38 & \textbf{0.2} & 0.2 & 0.17 & 0.1 & 0.07 \\
Modifying clauses & 0.69 & 0.10 & 0.62  & 0.52 & \textbf{0.66} & 0.62 \\
Modifying phrases* & -- & -- & -- & -- & -- & -- \\
Mood & 1.0 & \textbf{0.76} & 0.48 & 0.21 & 0.48 & 0.21 \\
New words & 0.84 & 0.16 & 0.08 & 0.16 & \textbf{0.4} & 0.36 \\
Parallelism & 0.95 & 0.1 & 0.81 & 0.81 & 0.84 & \textbf{0.84} \\
Phrases b.o. nouns & 1.0 & 0.03 & \textbf{0.97} & 0.74 & 0.84 & 0.9 \\
Phrases b.o. verbs & 0.72 & 0.0 & 0.67 & 0.8 & 0.67 & \textbf{0.97} \\
Predication & 0.9 & 0.13 & 0.4 & 0.17 & 0.2 & \textbf{0.57} \\
Prosody \& punct.** & -- & -- & -- & -- & -- & -- \\
Sentence arch. & 0.97 & 0.0 & 0.73 & 0.53 & \textbf{0.73} & 0.6 \\
Series & 1.0 & 0.0 & 0.83 & 0.87 & \textbf{0.96} & 0.83 \\
Subject choices & 0.87 & 0.13 & 0.23 & \textbf{0.5} & 0.47 & 0.1 \\
Tense & 0.73 & 0.1 & 0.4 & 0.5 & 0\textbf{.63} & 0.6 \\
Tropes & 0.7 & 0.09 & 0.31 & 0.16 & 0.44 & \textbf{0.47} \\
Verb choices & 0.73 & 0.23 & 0.2 & 0.27 & 0.53 & \textbf{0.77} \\
\midrule
\bf All features & \bf 0.83 & \bf 0.11 & \bf 0.56 & \bf 0.39 & \bf 0.51 & \bf 0.53 \\
\bottomrule
\end{tabular}
\end{center}
\caption{\label{table:results} Accuracy of annotations where all annotators agree. The highest scores, excluding human annotators, are highlighted in bold font. In GPT Version 1 (V1) the prompts to GPT include all properties of a given feature and we ask GPT to choose zero or more applicable properties. In GPT Version 2 (V2) columns, each property gets its own prompt, and GPT is asked to provide a yes/no answer as to whether the property is applicable. The FT column contains scores, where GPT-3.5 was fine-tuned on the majority, vote human annotations, and each property receives its own prompt. Features marked with an asterisk are extracted using existing NLP techniques and do not need to be annotated. Features marked with double asterisks are too sparse to be useful. }
\end{table}


\section{Discussion}\label{results}
We aim to make the best of the noisy data obtained through human expert annotations by utilizing the state-of-the-art large language models, which have recently dominated NLP research. Identifying linguistic features and rhetorical devices in natural language is not trivial for humans, as is borne out of the low inter-annotator agreement on our task. However, it seems this task would be well suited to an LLM for the following reasons: it does not require reasoning, it does not require any highly specialized domain knowledge that wouldn't have been available in the data the LLM was trained on, nor does it require access to facts or events which would have occurred post the training date. The one thing that LLMs are particularly good at, pattern recognition and general `knowledge' (arising from the vast amounts of publicly available data they were trained on) would suggest that they are ideally suited to our task. This aligns with the generally accepted premise that people are good at logical reasoning while machines are good at finding patterns in large volumes of data. 

Our research highlights another interesting, perhaps less conspicuous, difference between human annotators and GPT. What we found was that expert human inter-annotator agreement is low when it comes to non-trivial linguistic feature identification tasks. Conversely, GPT, when called with a low-temperature setting, achieves consistency levels around 80\% (see Figure \ref{fig:gpt-vs-human}). However, when humans do agree, they are correct, overall, 83\% of the time, while the maximum GPT accuracy we were able to achieve is only 56\% (see Table \ref{table:results}). This suggests that we can confidently rely on the established majority voting mechanism, or `wisdom of the crowd' approach, when it comes to traditional human annotations, but not when it comes to GPT.

This brings us to our next point, which is that every setting appears to demand its own exploration, and the techniques that work in one scenario do not always translate to another. In particular, we found that providing a small set of carefully crafted high-quality data points for fine-tuning does not necessarily increase the accuracy of GPT-3.5, as discussed in the previous section. In addition to different settings demanding different approaches, on a more granular level, not all features respond similarly to the same techniques. For example, `Mood' and `Emphasis' accuracy was highest using the original prompts, which performed poorly for all the other features. The `Aspect', `New words', and `Subject choices' features are relatively easy for humans but particularly difficult for GPT. And fine-tuned GPT outperforms humans on `Phrases built on nouns'.  These insights enable us to make cost/benefit decisions about utilizing human annotators vs GPT, prompting technique, or fine-tuning. 

We have run our experiments on GPT-3.5 and GPT-4 for comparison purposes, and no doubt the next version of GPT will be even better. However, our goal is to find ways to optimize not only for accuracy but also for cost. Our fine-tuned (FT) models perform on par with GPT-4 at 10x less the price. Since we intend to annotate the PTC corpus, which comprises approx. 20,000 sentences, we can estimate that using our FT models would cost approx. \$2,000\footnote{This is a rough estimate based on our spending to date. A more precise estimate can be calculated by estimating the number of tokens input and output to and from GPT.}, GPT-4 would cost \$20,000, and human annotators would cost \$90,000 (at \$1.50 per sentence, per expert, using three experts). These orders of magnitude differences can be critical to a researcher's budget and ability to proceed.

\section{Related Work}

Since the release of ChatGPT in November 2022, researchers \cite{Gilardi_Alizadeh_Kubli_2023, He_Lin_Gong_Jin_Zhang_Lin_Jiao_Yiu_Duan_Chen_2023,Yu_Li_Su_Fuoli_2023, Pangakis_Wolken_Fasching_2023,Tornberg_2023, Kuzman_Mozetič_Ljubešić_2023, Thapa_Naseem_Nasim_2023,Ding_Qin_Liu_Chia_Joty_Li_Bing_2023, Bansal_Sharma_2023,Feng_Narayanan_2023,Alizadeh_Kubli_Samei_Dehghani_Bermeo_Korobeynikova_Gilardi_2023,Savelka_2023,huang_ahte_speech_2023,Tekumalla_Banda_2023,Zhang_Li_Ma_Zhou_Zou_2023} have explored using it as a text annotation tool to reduce reliance on expensive human annotated data for downstream NLP tasks. Various prompt engineering strategies have been proposed including zero-shot, few-shot, chain-of-thought (CoT), and active learning. 


\citet{Wang_Liu_Xu_Zhu_Zeng_2021} first explore using GPT-3 (the model upon which ChatGPT was built) for text annotation and show that data labelled by GPT-3 achieves the same performance on several NLP and NLU tasks as human labelled data at a fraction of the cost. Furthermore, models trained using the GPT-3 labelled data outperform GPT-3 on their respective tasks. However, the authors suggest that labeling by GPT-3 should only be used in low stakes scenarios due to its now well known tendencies to ``hallucinate'' \cite{Ji_2023_hallucination_survey}, and/or perpetuate undesirable biases. \citet{Reiss_2023} also cautions against using ChatGPT-3.5 for zero-shot annotation tasks, pointing out the intra-coder inconsistencies (e.g. the inconsistencies among ChatGPT-3.5 responses given identical prompts). To improve the accuracy of ChatGPT annotations, \citet{Reiss_2023} recommends using majority voting based on three, or ideally more, repetitions of each single input. We show that this does not improve results in our setting. Taking this a step further, \citet{Davani_Díaz_Prabhakaran_2022} show that taking a majority vote before training is not as effective as training a multi-task model using all annotators (ChatGPT repetitions) and subsequently predicting the final label as an aggregation of each task output. Similar to \citet{Reiss_2023}, \citet{Pangakis_Wolken_Fasching_2023} point out that while ``LLM performance for text annotation is promising, it is highly contingent on both the data set and the type of annotation task, which reinforces the necessity to validate on a task-by-task basis.'' The authors propose an iterative strategy that evaluates the LLM annotations on a small subset of human expert annotated data until acceptable performance is reached. We follow a similar strategy.


Despite these potential pitfalls, others have shown promising results. \citet{Kuzman_Mozetič_Ljubešić_2023} compare a fine-tuned XLM-RoBERTa model against ChatGPT-3 for genre identification, a task deemed difficult for human annotators, where the inter-annotator agreement is low. While this is not strictly an annotation comparison, we can think of the GPT zero-shot classifier as an annotator where GPT achieved an accuracy score of 72\% against the human-annotated test set. GPT outperformed the fine-tuned model by 5 points on an out-of-distribution data set given a similar task. In contrast, \citet{He_Lin_Gong_Jin_Zhang_Lin_Jiao_Yiu_Duan_Chen_2023} propose a method which surpasses human annotators on some tasks and reaches near parity on others. Unlike the zero-shot scenario, the proposed method involves few-shot chain-of-thought (CoT)  \cite{Wei_Wang_Schuurmans_Bosma_Ichter_Xia_Chi_Le_Zhou_2023} prompting. Uniquely, GPT-3.5 is used to generate the explanations, which are then used as few-shot examples in the final prompt. Having GPT generate the explanations is justified by attributing an emergent reasoning capability to GPT. In contrast, we provide expertly written definitions and real-world examples in our prompts. However, when we examine the GPT-3.5 explanations in our setting, we cannot identify any emergent reasoning capabilities - see section \ref{promt-engineering} for details. 

\citet{Alizadeh_Kubli_Samei_Dehghani_Bermeo_Korobeynikova_Gilardi_2023} extend the research into LLM annotation beyond ChatGPT to open-source (OS) models. They show that while ChatGPT generally outperforms human annotators and OS models on most tasks, OS models outperform ChatGPT on some tasks and approach ChatGPT performance on others. OS models also outperform human annotators at a lower cost, with all the open-source benefits such as transparency, reproducibility, and data privacy.

\citet{Bansal_Sharma_2023} take an active learning approach, where the LLM is utilized to annotate the most informative inputs based on a novel conditional informativeness sampling strategy, thereby reducing the cost of annotation even further while improving the performance of downstream NLP models trained on the LLM annotated data.

It's clear from the literature that the ideal prompt engineering methods and hyper-parameters vary across tasks, especially when those tasks are complex or particularly difficult for human annotators. Our work focuses on the scenario where the task is unique and complex, and human annotator agreement is low, and therefore
merits its own study.

\section{Conclusion}\label{conclusion}

In this study, we have identified linguistic features based on the premise that propaganda techniques can be decomposed into rhetorical devices and linguistic characteristics, as seen in the use of language for persuasion. Given our limited budget, we employed three expert annotators to label 350 sentences from the PTC corpus, to the best of our knowledge, the most granular propaganda technique corpus available. With the help of our small human-annotated data set, we investigated whether a large language model like GPT can be utilized to annotate the remaining \~20,000 sentences. The impetus for our experiments and solutions is driven by an acutely limited budget, thus the solutions we propose are necessarily a trade-off between accuracy and cost. 

In addition, we find certain unexpected differences between human annotators and GPT. In particular, we discovered that majority voting of GPT responses does not improve accuracy, and fine-tuning on a small but high-quality data-set is also not an effective strategy. Both of these findings are contrary to usual practices in machine learning, and we must emphasize again that each setting demands its own exploration. 

Finally, we fine-tune GPT-3.5 using the human-annotated data set by generating examples for each feature/property combination by taking a majority vote for the positive examples and the absence of a given label from all annotators as negative examples. Even with this noisy data, we achieve an accuracy on par with GPT-4 at 10x less the cost. We make all our code, including our web-based annotation application, the properties and their definitions and examples, and the GPT prompts available in GitHub.\footnote{\url{https://github.com/kyleiwaniec/rhetoric-annotation}}

\section{Limitations and Future Work}\label{limitations}

Our biggest limitation is our budget. However, even without this constraint, the task is not trivial, even for expert annotators. Majority voting by itself still leads to a lot of noise, and we intend to perform several more iterations on the 350 sentences with our human annotators to reach a higher agreement. We performed our experiments on GPT as it is the subject of considerable attention. In the future, we plan to test other models. In particular, LLaMA \cite{touvron2023llama} and Mistral \cite{jiang2023mistral} are open source and available in different sizes, making them especially suitable for analyzing the trade-off between cost and accuracy. 

Furthermore, the feature set we created is based on literature, but this, too, could benefit from additional research. We have yet to see if these features improve the detection of propaganda techniques. While this is outside the scope of this paper, it is the next natural step. By making these available, we hope to aid other researchers in pursuing the goal of interpretable propaganda detection.

\begin{acks}
This publication has emanated from research conducted with the financial support of Science Foundation Ireland under Grant number 18/CRT/6183. For the purpose of Open Access, the author has applied a CC BY public copyright licence to any Author Accepted Manuscript version arising from this submission.
\end{acks}

\bibliographystyle{ACM-Reference-Format}
\balance
\bibliography{base}

\appendix
\section{Features}

\begin{table*}[t]
\small
\centering
\begin{tabular}{p{4.5cm}@{\hskip .5cm}p{10.6cm}}

\multicolumn{1}{l}{{\bf Features}} & \multicolumn{1}{l}{{\bf Properties}} \\  
\toprule
\\
\bf Word choices & \\ 
\midrule
\\
Figures of word choice & ['agnominatio', 'metaplasms', 'polyptoton', 'ploce', 'anatanaclasis', 'synonyms', 'rhetorical conditional', 'correctio', 'emphasis'] \\
Language of origin & ['Old English Core', 'Norman French', 'Latin/Greek'] \\
Language varieties & ['correctness', 'clarity', 'forcefulness', 'low', 'middle', 'high', 'dialects/registers', 'register shift', 'cliches/idioms', 'maxims/proverbs', 'allusions'] \\
Lexical and semantic fields & ['prototype', 'abstract', 'concrete', 'indefinite/thesis', 'definite/hypothesis'] \\
New words and changing uses & ['foreign borrowing', 'compounds', 'prefixes/suffixes', 'clipping', 'blends', 'conversions', 'catachresis', 'acronyms', 'proper nouns to common nouns', 'analogy', 'fabrication', 'onomatopoeia', 'taboo deformation', 'doubling'] \\
Tropes & ['synecdoche', 'metonymy', 'antonomasia', 'metaphor', 'allegory', 'simile', 'full analogies', 'irony', 'hyperbole', 'litotes', 'amphiboly', 'paradox', 'paralepsis/praeteritio', 'euphemism'] \\
\\
\bf Sentences & \\ 
\midrule
\\
Aspect & [`simple', `perfect', `progressive', `perfect progressive'] \\
Emphasis & [`by position', `by sentence role', `from inversions', `combinations'] \\
Figures of argument & [`strategic repetition', `antithesis ', `antimetabole', `defintion as figure', `induction', `eduction'] \\
Modifying clauses & [`subordinate', `conditional', `comparitive', `adjective', `noun'] \\
Modifying phrases & [`prepositional phrases', `single word modifiers', `multiplying and embedding modifiers'] \\
Mood & [`indicative', `subjunctive', `exclamatory', `interrogative', `imperative', `optative'] \\
Parallelism & [`in syllables (isocolon)', `in stress patterns (iambic parameter)', `grammatical structure (parison)', `comparison (parallelism in argument)'] \\
Phrases built on nouns & [`appositives', `absolute construction', `resumptive modifier', `summative modifier'] \\
Phrases built on verbs & [`participal phrases', `inifinitive phrases'] \\
Predication & [`active', `stative', `compound'] \\
Prosody and punctuation & [`innotation', `stress', `rhythm'] \\
Sentence architecture & [`left branching', `middle branching', `right branching', `Periodic sentences', `loose sentences'] \\
Series & [`bracketing', `order (incrementum)', `gradatio', `polysyndenton', `asyndenton'] \\
Subject choices & [`humans', `rhetorical participants', `things', `abstractions', `concepts', `slot fillers'] \\
Tense & [`present', `past', `future', `progression'] \\
Verb choices & [`negation', `modality', `nominalization', `personification', `multiplication'] \\
\bottomrule
\end{tabular}
\caption{\label{table:features} Features and their properties extracted from \textit{Rhetorical Style: The Uses of Language in Persuasion} \cite{fahnestock2011rhetorical}. 
  }
\end{table*}

\end{document}